\documentclass{ifacconf}
\usepackage{enumitem}
\usepackage{amsmath,amssymb,amsfonts}
\usepackage{algorithmic}
\usepackage{textcomp}
\usepackage{xcolor}
\usepackage{multirow}
\usepackage{mathtools}
\usepackage{graphicx}      
\usepackage{natbib}        
\begin{document}
\begin{frontmatter}

\title{Facilitating Cooperative and Distributed Multi-Vehicle Lane Change Maneuvers}


\author[First]{Hansung Kim} 
\author[First]{Francesco Borrelli} 

\address[First]{Department of Mechanical Engineering, University of California-Berkeley, 
  Berkeley, CA 94703 USA (e-mail: hansung@berkeley.edu,~fborrelli@berkeley.edu)}

\begin{abstract}                
A distributed coordination method for solving multi-vehicle lane changes for connected autonomous vehicles (CAVs) is presented. Existing approaches to multi-vehicle lane changes are passive and opportunistic as they are implemented only when the environment allows it. The novel approach of this paper relies on the role of a \textit{facilitator} assigned to a CAV. The \textit{facilitator} interacts with and modifies the environment to enable lane changes of other CAVs. Distributed MPC path planners and a distributed coordination algorithm are used to control the \textit{facilitator} and other CAVs in a proactive and cooperative way. We demonstrate the effectiveness of the proposed approach through numerical simulations. In particular, we show enhanced feasibility of a multi-CAV lane change in comparison to the simultaneous multi-CAV lane change approach in various traffic conditions generated by using a data-set from real-traffic scenarios.
\end{abstract}


\end{frontmatter}

\section{Introduction}
\vspace{-0.2cm}
Connected vehicle technology including vehicle-to-vehicle (V2V), vehicle-to-cloud (V2C), and vehicle-to-infrastructure (V2I) communication reduces the risk of collisions and energy usage, enabled by the enhanced knowledge of the environment \citep{v2v}. By leveraging connectivity, connected autonomous vehicles (CAVs) can form platoons with short headway time and coordinate multi-vehicle maneuvers such as lane change \citep{GUANETTI201818}. Platooning increases traffic throughput by allowing small inter-vehicle distances, reducing motion delays in response to changing conditions,  and enhances energy efficiency by  minimizing undesirable braking using shared motion forecast. Because of these advantages, there has been an increasing interest in vehicle platooning and its cooperative systems to enhance safety, energy efficiency, robustness of CAV controllers in the last decade. For instance, \cite{eco-adaptive} demonstrated energy savings of 12\% and jerk reduction of 31\% using a model predictive control architecture to incorporate preceding platoon vehicle's state and acceleration forecast information received via V2V in a model predictive control (MPC) framework while guaranteeing collision avoidance.
\begin{figure*}[t!]
    \centering \includegraphics[width=0.6\textwidth]{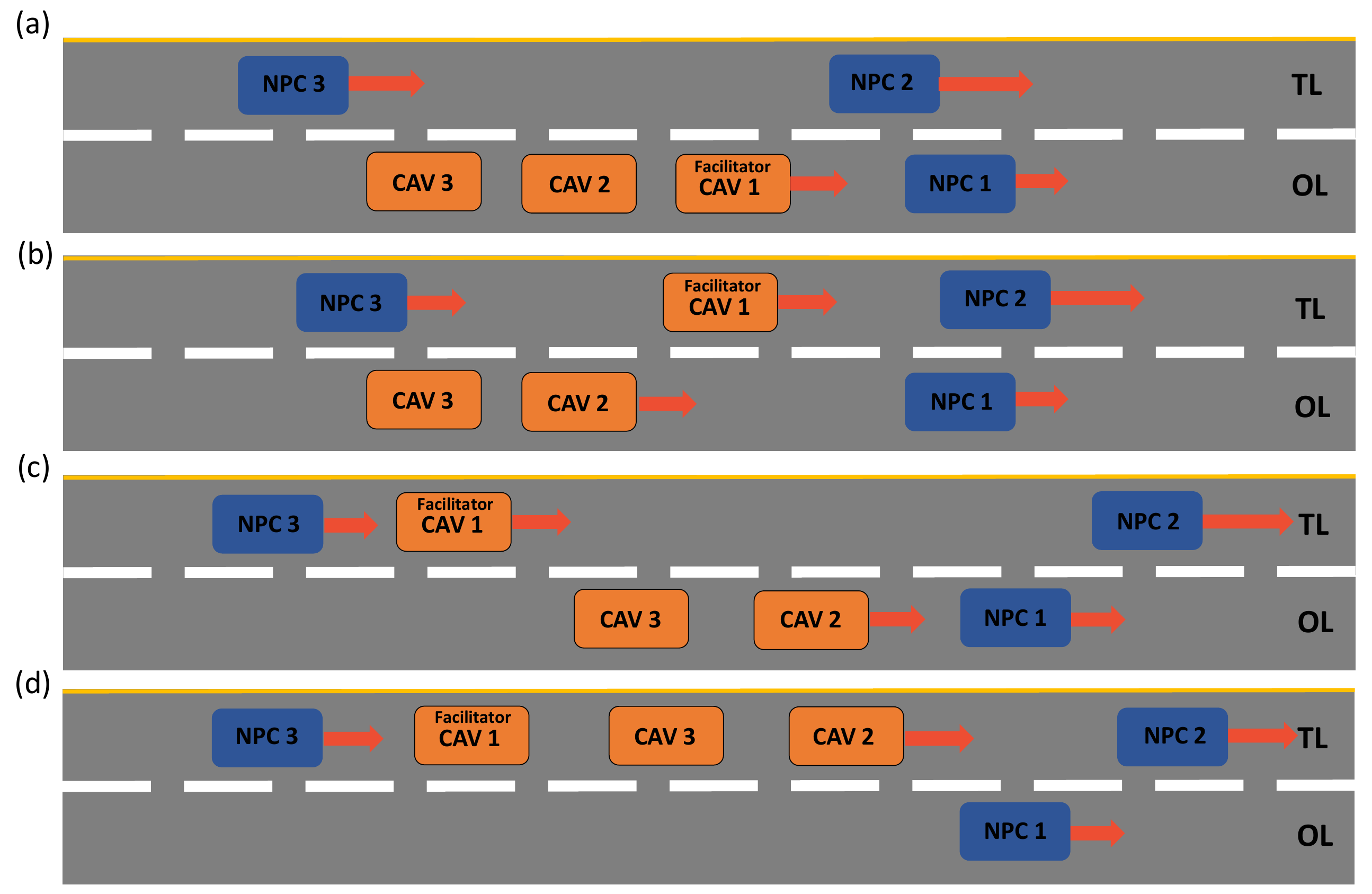}
    \caption{Snapshots of the proactive and cooperative lane change strategy: (a) the facilitator change lane into the Target Lane (TL), (b) CAV 1 and CAV 2 accelerate while the facilitator decelerate to create free space for CAV 1 and CAV 2, (c) lane change for CAV 1 and CAV 2 is rendered feasible, (d) CAV 1 and CAV 2 completes a lane change}
    \label{fig:scenarios}
\end{figure*} 
With advancements in vehicle connectivity technology, more complex coordinated and cooperative maneuvers such as platoon merging, splitting, reconfiguration, and lane change can be implemented. In existing literature, cooperative platoon maneuver coordination is typically formulated in two ways: 1) a centralized control assuming either no uncooperative non-player character (NPC) vehicles are present in the environment or their state forecasts are exactly known \citep{sun_horowitz_tan_2003,multi-vehicle,roya}, 2) a distributed control assuming no NPC vehicles are present in the environment \citep{decentr_no_npc,beyond}. However, NPC vehicles, typically human-driven vehicles, are present in the environment and their motion forecasts are uncertain in a real world environment. Moreover, platoon lane change maneuvers in \citet{sun_horowitz_tan_2003}, \citet{multi-vehicle}, and \citet{roya} are executed simultaneously and are \textit{passive and opportunistic} as the platoon adapt their policies to the changing environment. When the environment is configured such that a platoon lane change is not possible, the platoon remains in its original lane until the environment evolves into a configuration where safe execution of a lane change is possible. 
Simultaneous platoon lane change has fewer opportunities to change lanes in certain traffic conditions than a single vehicle because larger free space in the target lane is required for multiple vehicles for a safe execution of a lane change maneuver. These limitations underscore the increased complexity of lane change maneuvers when considering a multi-vehicle platoon interacting with surrounding vehicles.     

In this paper, we propose a distributed coordination method to implement a novel \textit{proactive and cooperative} multi-CAV platoon lane change strategy. In the proposed strategy, a CAV performs a role of a \textit{facilitator}. The \textit{facilitator} actively interacts and modifies the environment to aid the multi-CAV platoon in executing a lane change while reducing the interaction between the multi-CAV platoon and NPC vehicles. A multi-CAV platoon refers to a single-lane platoon with a small number of interconnected vehicles. 
Our approach to distributed coordination uses three \textit{modes} (\textit{Lane Keeping}, \textit{Lane Change}, \textit{Gap Regulation}) to describe the CAV maneuver modes. Finite state machines which control the mode transitions of each CAVs are also used to coordinate CAV maneuvers. The mode transition logic depends on not only the state of the controlled CAV but also on those of other CAVs in the multi-CAV platoon by design. In each mode, distinct constraints and cost functions in distributed MPC path-planners are designed to execute the desired maneuvers for multi-CAV coordination.

We demonstrate that the proposed lane change strategy enhances feasibility of multi-CAV platoon lane change compared to the \textit{passive and opportunistic} multi-CAV platoon lane change strategy through numerical simulation. In numerical simulation, a real-world NGSIM dataset \citep{ngsim_data} is used to generate realistic initial conditions for the set of CAVs and NPC under consideration. 
The contributions are summarized as follows:

\begin{enumerate}
    \item A novel proactive and cooperative multi-CAV platoon lane change strategy which relies on the role of a \textit{facilitator}.
    \item A distributed coordination method consisting of a MPC based path-planner with three modes---in which different terminal constraints and safety constraints as well as cost functions are used---and a finite state machine that controls the mode transitions of the CAVs.
    \item A simulation study demonstrating enhanced feasibility of multi-CAV platoon lane change using the proposed strategy compared to opportunistic simultaneous multi-CAV lane change strategy. 
\end{enumerate}

The remainder of the paper is organized as follows. Section \ref{facilitator} introduces the concept of a \textit{facilitator} and its role in \textit{proactive and cooperative} multi-CAV platoon lane change strategy. Section \ref{models} discusses the vehicle dynamic model and NPC prediction model used in our MPC path-planner formulation. Section \ref{dist_coord} discusses the distributed coordination method including the higher-level FSM and distributed MPC path-planner used to implement the proposed lane change strategy. The numerical simulation setup and results of multi-vehicle lane change in dense traffic is presented in Section \ref{numerical_sim}. Finally, Section \ref{Conclusion} concludes the paper and discusses future research directions.

\section{Proactive and Cooperative multi-CAV platoon Lane Change Strategy: The Concept of Facilitator} \label{facilitator}

Consider an environment with a multi-CAV platoon of $n$-CAVs in the same lane as shown in Fig. \ref{fig:scenarios}(a). Assuming that the multi-CAVs' objective is to change into the Target Lane (TL) from the Original Lane (OL), the multi-CAV platoon must interact with surrounding vehicles when executing the maneuver. For instance, the multi-CAV platoon (orange vehicles) of 3 CAVs in the environment shown in Fig. \ref{fig:scenarios}(a) must consider NPC 2 and NPC 3's positions and velocities to determine the feasibility of a lane change. Using the \textit{passive and opportunistic} lane change strategy, the multi-CAV platoon may change lanes only when the surrounding vehicles' positions and velocities are configured in a way such that lane change is simultaneously feasible for all CAVs in the multi-CAV platoon (i.e. if lane change is feasible for all but one vehicle such as CAV 3, the multi-CAV platoon does not change lanes). In certain traffic conditions, the multi-CAV platoon lane change may be infeasible for a long time. However, using our proposed lane change strategy, the \textit{facilitator} (without loss of generality, CAV 1 will be assigned the facilitator role in this work) enables the lane change of the other CAVs in the multi-CAV platoon by \textit{proactively} manipulating the environment in favor of the multi-CAV platoon. For instance, the facilitator changes lanes when feasible and regulates the free space between itself and the vehicle in front of it (NPC 2) as shown in Fig. \ref{fig:scenarios}(b)-(c). This allows CAV 2 and CAV 3 to change lanes into the free space regulated by the facilitator without considering \textit{NPC 3}. Thus, the complexity of interaction between surrounding vehicles and the multi-CAV platoon is reduced.  


In order to implement the \textit{proactive and cooperative} lane change strategy, a sequence of coordinated maneuvers must be executed by each CAVs in the multi-CAV platoon as shown in Fig. \ref{fig:scenarios}(a)-(d). 

\
\section{Vehicle Models} \label{models}

In this section, we first describe the kinematic bicycle model of nonlinear vehicle dynamics in Frenet frame. Then, we describe the NPC prediction model used in our path-planner formulation. In this work, we assume the OL center line and curvature information are known.

\subsection {Vehicle Dynamics}

\begin{figure}[h!]
\centering
  \includegraphics[width=0.79\linewidth]{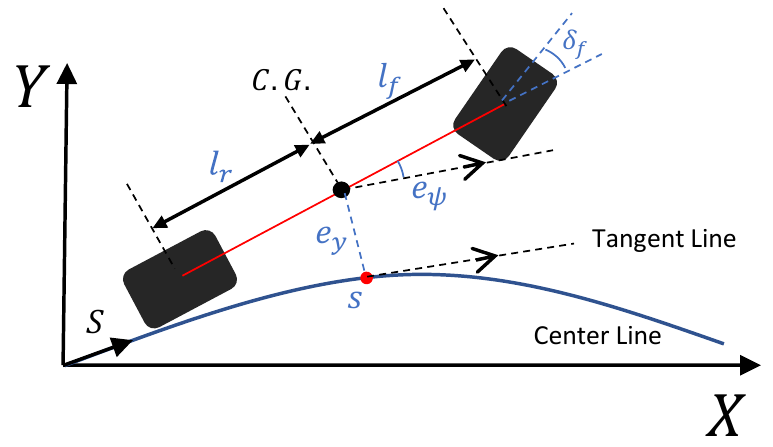}
  \caption{The kinematic bicycle model in Frenet Frame}
  \label{fig:kinematic}
\end{figure}

The $i$-th CAV state variables are defined as $\mathbf{z}^i(t) := [s^i(t),e_y^i(t),e_\psi^i(t),v^i(t)]^\top$ for $i\in\{1,2,...,n\}$, where $s^i(t)$ and $e_{y}^i(t)$ represent the longitudinal and lateral displacement of the vehicle with respect to the desired lane's center line, respectively, as seen in Fig. \ref{fig:kinematic}. $n$ is the number of CAVs in the platoon. $e_\psi^i(t)$ is the heading angle difference between that of the vehicle and the center line and $v^i$ is the longitudinal velocity of the vehicle at the center of gravity (C.G.). The control input vector is $\mathbf{u}^i(t)=[a^i(t),\delta_f^i(t)]^\top$, where $a^i(t)$ is the longitudinal acceleration of the vehicle at the C.G. and $\delta_f^i(t)$ is the steering angle. The vehicle dynamics are 
\begin{equation} \label{kinematic_bicycle}
\begin{split}
\dot{s}^i(t) & = v^i(t) \, \tfrac{cos\,e_\psi^i(t)}{1-e_y^i(t)\, \kappa(t)}, \\
\dot{e_y}^i(t) & = v^i(t) \, sin\,e_\psi^i(t), \\ 
\dot{e_\psi}^i(t) & = v^i(t) \, (\tfrac{tan\,\delta_f^i(t)}{l_f^i + l_r^i}-\tfrac{\kappa(t) \, cos\,e_\psi^i(t)}{1-\kappa(t) \, e_y^i(t)}),\\
\dot{v}^i(t) & = a^i(t), \\
\end{split}
\end{equation}
$l_f^i$ and $l_r^i$ are distance from the front axle to the C.G. and distance from the rear axle to the C.G., respectively. $\kappa(t)$ is the curvature of the reference centerline. 

The continuous-time dynamics model is discretized using forward Euler method with sampling time $\Delta t$ as follows
\begin{equation} \label{kinematic_bicycle_dt}
\mathbf{z}^i(k+1)  = f(\mathbf{z}^i(k),\mathbf{u}^i(k)) 
\end{equation}
where $\mathbf{z}^i(k)$ denotes the state vector $\mathbf{z}^i$ at time $k\Delta t + t$ for $k \in \mathbb{Z}^+_0$.

\subsection{NPC Prediction Model}
During the prediction horizon $N$, we use a longitudinal uncertain prediction model to predict NPC vehicle's motion. The model assumes uncertainty on acceleration and that the NPC vehicle does not change lanes within the prediction horizon. Given the OL centerline and road curvature information: $(\kappa_t,\dots,\kappa_{t+N})$, the resulting discretized dynamics of the NPC vehicle is as follows 
\begin{equation} \label{npc_pred_model}
\begin{split}
s^{NPC}_{k+1|t} & = s^{NPC}_{k|t} + \Delta t\, v^{NPC}_{k|t} \, \tfrac{cos\,e^{NPC}_{\psi, k|t}}{1-e^{NPC}_{y, k|t}\, \kappa_{k}}, \\
e^{NPC}_{y, k+1|t} & = e_y^{NPC}(t), \\ 
e^{NPC}_{\psi, k+1|t} & = 0,\\
v^{NPC}_{k+1|t} & = v^{NPC}_{k|t} + \Delta t \; w_k, \\
\end{split}
\end{equation}
for $k = t, ... , t+N$; where $e_y^{NPC}(t)$ is $e_y$ of the NPC vehicle at time $t$. ${w_k} \sim N(0, \sigma^2)$ is a random variable representing acceleration at time $k$. The variance $\sigma^2$ is the design parameter of the model and depends on the acceleration limitation of the NPC vehicle. In our simulation, we set $\sigma^2=0.2$ $m/s^2$. The proposed approach can accommodate different prediction model for improved prediction accuracy and robustness.


\section{Distributed Coordination} \label{dist_coord}

A distributed coordination of CAVs to implement a proactive and cooperative lane change strategy is designed as follows: MPC Path-planners with three modes in which different terminal constraints and safety constraints as well as cost functions are used to plan trajectories of each CAVs. Each CAVs have corresponding finite state machines that manage the mode transitions of the planners and coordinate multi-CAV platoon maneuvers. In this section, the FSM and MPC path-planner are formulated. Furthermore, we discuss methods to avoid trajectory conflicts between CAVs in distributed path planning of a multi-vehicle system.

\subsection{Finite State Machine}
The CAVs in the platoon are indexed by $i$ starting from the front-most CAV. Note that the facilitator is always assigned $i=1$. 
The FSM makes use of parameters and binary inputs associated with every CAVs in the platoon and denoted by the superscript $i\in\{1,2,...,n\}$ for the $i$-th CAV. 
The parameters and binary inputs used in our proposed FSM are listed in Table \ref{tab:FSM}. 

\begin{table} [h!]
\caption{FSM parameters and inputs} \label{tab:FSM}
\centering
\begin{tabular}{c|c }
\hline
\textbf{Parameter} & \textbf{Definition} \\ 
\hline
$TL$ & Target lane for lane change\\
$OL$ & Original lane of the platoon\\
$\Delta s_{TL}$ & Distance from the facilitator \\ &to the preceding vehicle in the $TL$\\
$\Delta s_{OL}$ & Distance from the facilitator \\ &to the preceding vehicle in the $OL$\\
$d^1$ & $\min(\Delta s_{TL},\Delta s_{OL})$ \\ 
$d^{i}$ & Signed distance from $i$-th CAV \\ & to the facilitator, for $i:[2,n]$ \\ 
$d^{1}_{des}$ & Desired minimum distance from the \\ & facilitator to the preceding \\ & vehicles in $TL$ or $OL$ \\ 
$d^{i}_{des}$ & Desired minimum signed distance \\ & from $i$-th CAV to the facilitator\\
& for $i:[2,n]$\\
\hline
\textbf{FSM binary inputs} & \textbf{Definition} \\
\hline
$LC_{feasible}^{i}$ & Lane change feasibility of the \\ & $i$-th CAV in the platoon  \\
$LC_{complete}^{i}$ & Lane change completion of the \\ & $i$-th CAV in the platoon \\
$LC_{complete}^{P}$ & $\prod_{i=2}^{n}LC_{complete}^{i}$ \\
$r_{satisfied}^{i}$ & $d^{i} \geq d^{1}_{des} $ satisfied \\ 
\hline
\end{tabular}
\end{table}

The desired minimum longitudinal inter-vehicular distances are defined as the following:
\begin{equation}
\label{eq:df}
d^1_{des} = n\cdot d_{safe} 
\end{equation}
\begin{equation} \label{eq:Pi}
d^{i}_{des} = (n-i)\cdot d_{safe}, \; \text{for} \; i = 2, 3, ... , n 
\end{equation}
where $d_{safe}$ is the user-defined safe vehicular distance and impacts the conservativeness of the controlled vehicle's driving behavior. In the context of multi-CAV lane change, $d^1_{des}$ represents the desired free space that the facilitator must create. $d^{i}_{des}$ corresponds to the desired distance (relative to the facilitator) for the $i$-th CAV to execute a lane change into the free space created by the facilitator. These inter-vehicular distances fully describe the relative positions of the multi-CAV platoon and the facilitator. 

\begin{figure*}[t!]
    \centering \includegraphics[width=0.65\linewidth]{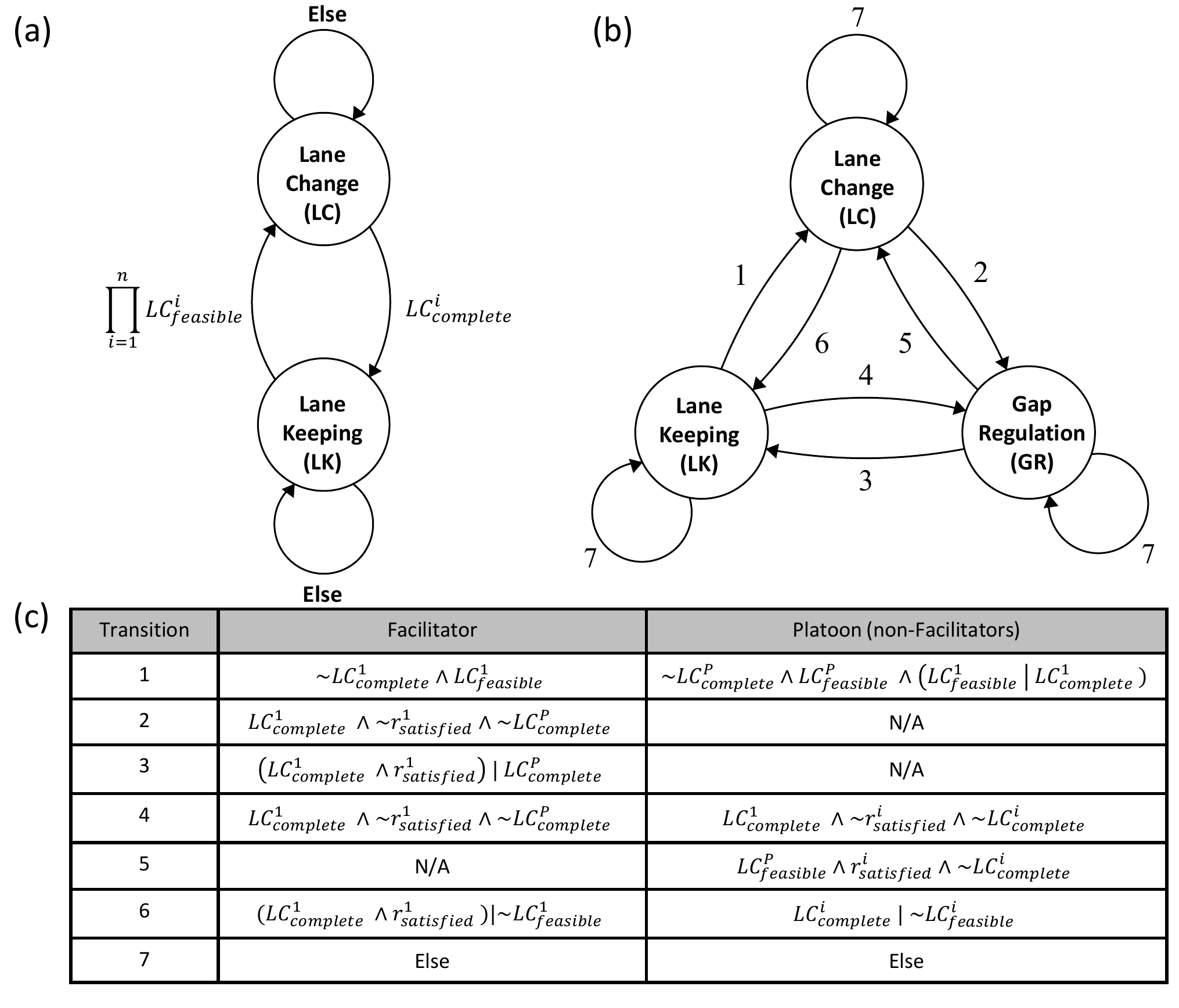}
    \caption{Finite state machines of path planner modes in a highway driving scenario: (a) a FSM representing a passive and opportunistic lane change for a platoon P, (b) FSM with an additional mode---\textit{Gap Regulation}---and its switching conditions listed in (c) for the proposed proactive and cooperative lane change strategy. Refer to Table \ref{tab:FSM} for definitions of the mode switching conditions.    
    }
    \label{fig:FSM}
\end{figure*}

Typically, the simplest descriptions of highway driving of an autonomous vehicle is categorized into two modes: \textit{Lane Keeping} (LK) and \textit{Lane Change} (LC). In lane keeping mode, a CAV stays in its current lane and attempts to follow the reference velocity while maintaining a safe distance from the preceding vehicle. In lane change mode, the CAV attempts to change lanes into TL if such maneuver is safe and feasible. An FSM that shows the transition map between the two modes for a multi-CAV platoon is shown in Fig. \ref{fig:FSM}(a). Note that this two-mode FSM is sufficient for a \textit{passive and opportunistic} multi-CAV platoon lane change strategy, in which switch to \textit{Lane Change} mode occurs if and only if a platoon-wise lane change is simultaneously feasible ($\prod_{i=1}^{n} LC_{feasible}^i$ is \textit{True}). After completing a multi-CAV platoon lane change, the planner switches back to \textit{Lane Keeping} mode.

Our proposed FSM has an additional mode: \textit{Gap Regulation} (GR) as shown in Fig. \ref{fig:FSM}(b). In \textit{Gap Regulation} mode, the CAV regulates the inter-vehicular distance of interest (gap) to the desired longitudinal distance. By assigning different desired inter-vehicular distance per CAV, the multi-CAV platoon is regulated into a desired formation in a distributed manner. Furthermore, the mode transition logic is different for the \textit{facilitator} and the multi-CAV platoon but not independent. The mode transition depends on not just the controlled CAV's states but also on the states of other CAVs as a mechanism for coordination. For instance, the multi-CAV platoon (excluding the facilitator) switches to \textit{Gap Regulation} mode only after the facilitator has completed a lane change into the TL. In this mode, the \textit{facilitator} modifies the environment and creates free space in the TL for the rest of the multi-CAV platoon to change lanes while the rest of the multi-CAV platoon gets in desired formation to execute a simultaneous lane change as shown in Fig. \ref{fig:scenarios}(b)-(c). 

The mode transition logic for the facilitator and the non-facilitator CAV are listed in Fig. \ref{fig:FSM}(c) in separate columns. We denote the FSM mode of the $i$-th CAV at time t as $z^i_{FSM}(t)\in\{LK,LC,GR\}$ and the vector containing FSM binary inputs listed in Table \ref{tab:FSM} for all CAVs (i.e. $i=\{1,2,\cdots,n\}$) as $\mathbf{x}_{FSM}$. The $i$-th CAV's FSM mode transition at time t is denoted as  
\begin{equation}
\label{eq:FSM_dynamics}
z^i_{FSM}(t+\Delta t) = \Tilde{f}^i(z^i_{FSM}(t),\mathbf{x}_{FSM}),
\end{equation}
Note that the proposed FSM in Fig. \ref{fig:FSM} and Table \ref{tab:FSM} also allows simultaneous platoon-wise lane change when feasible by design. Further, in cases of infeasibilities while in \textit{LC} mode due to environmental factors, the mode transitions back to \textit{Lane Keeping} mode and returns to OL by design for safety.  

\subsection{MPC Path-planner}
Path planning is formulated as a constrained finite time optimization problem (CFTOP) to plan dynamically feasible, smooth, and collision-free trajectories for each vehicle. The MPC path-planner solves the CFTOP repeatedly in receding horizon at a constant frequency. The path-planner is designed using the MPC framework with an augmented state vector $\Tilde{\mathbf{z}}^i_{k|t} := {[\mathbf{z}_{k|t}^i{}^\top,d^i_{k|t}]}^\top$ where $\mathbf{z}^i_{k|t}$ is defined as $\mathbf{z}$ of the $i$-th vehicle at time $k$ predicted at time $t$. $d^i_{k|t}$ is $d^i$ at time $k$ predicted at time $t$ with the following dynamics
\begin{equation}
\label{eq:dk+1}
d^1_{k+1|t} = s^{NPC}_{k+1|t} - s^1_{k+1|t},
\end{equation}
\begin{equation} \label{eq:dk+1i}
d^{i}_{k+1|t} = s^1_{k+1|t}-s^i_{k+1|t}, \; \text{for} \; i \neq 1. 
\end{equation}
Therefore, the augmented state dynamics $\Tilde{\mathbf{z}}^i_{k+1|t}:=g(\Tilde{\mathbf{z}}^i_{k|t},\mathbf{u}^i_{k|t},w_k)$ is represented by (\ref{kinematic_bicycle_dt}) and $d^i_{k+1|t}$ which depends on the NPC prediction with uncertainty in (\ref{npc_pred_model}) when $i=1$. For $i\neq1$, $d^i_{k+1|t}$ is deterministic because it is the inter-vehicular distance between two CAVs. In a distributed multi-CAV system, the exact motion forecasts of other CAVs are communicated via V2V. Using the augmented state dynamics, the path-planner is defined as follows
\begin{subequations} \label{planner_cftoc}
\begin{flalign} 
\min_{\substack{\Tilde{\mathbf{z}}^i(\cdot|t),\\
\mathbf{u}^i(\cdot|t),\\ \epsilon^i(\cdot|t)}}&\sum_{k=t}^{t+N}\text{E}[\lVert Q(\Tilde{\mathbf{z}}^i_{k|t}-\Tilde{\mathbf{z}}^{i,des}_{k|t})\rVert^2_2] + c_{s}\cdot\epsilon^i_{k|t} + \nonumber && \\ 
&\sum_{k=t}^{t+N-1}\lVert\mathcal{R}_u\mathbf{u}^i_{k|t}\rVert^2_2 + \lVert\mathcal{R}_{\Delta\mathbf{u}}\Delta\mathbf{u}^i_{k|t}\rVert^2_2
\vspace{-0.25cm}
\end{flalign} 
\begin{align}
\text{s.t.}\quad
&\Tilde{\mathbf{z}}^i_{k+1|t} = g(\Tilde{\mathbf{z}}^i_{k|t},\mathbf{u}^i_{k|t},w_{k}), \label{p_dynamics}\\
&\Delta\mathbf{u}^{min}_{k|t} \leq \Delta\mathbf{u}^i_{k|t} \leq \Delta\mathbf{u}^{max}_{k|t}, \label{p_input_const}\\ &\mathbf{u}^i_{k|t}\in \mathcal{U}, \quad \forall k = t, ... ,t+N-1 \\
&\Tilde{\mathbf{z}}^i_{t|t} = \Tilde{\mathbf{z}}^i(t) \label{initial_cond} \\
&\mathbf{z}^i_{k|t}\in \mathcal{Z}, \quad \forall k = t, ... ,t+N \\
&\textbf{if }z^i_{FSM}(t)\in\{LC\}: \quad \nonumber \\ 
&\begin{aligned}
&\,\mathbf{z}^i_{t+N|t} \in \mathcal{Z}_f \label{p_lc_terminal}
&\end{aligned}\\
&\,\text{E}[s^{i, min}_{k|t}] + d_{safe}\leq s^i_{k|t} \leq \text{E}[s^{i, max}_{k|t}] - d_{safe} \label{p_lc_const}\\
&\textbf{if }z^i_{FSM}(t)\in\{LK,GR\}: \nonumber \\ 
&\, d_{safe}-\epsilon^i_{k|t} \leq \text{E}[s^i_{front, k|t}] - s^i_{k|t} \label{p_acc_const}\\
&\, \epsilon^i_{k|t} \geq 0, \quad \forall k = t, ... ,t+N \label{p_slack_const}
\end{align} 
\end{subequations}
where $\mathbf{u}^i(\cdot|t)=\{\mathbf{u}^i_{t|t}, ... , \mathbf{u}^i_{t+N-1|t}\}$ is the control input sequence over the horizon for the $i$-th vehicle, $N$. Furthermore, $\Delta \mathbf{u}^i_{k|t}:=\mathbf{u}^i_{k|t}-\mathbf{u}^i_{k-1|t}$ and $\Tilde{\mathbf{z}}^{i,des}_{k|t}$ is the desired state of the $i$-th vehicle which is determined by the lane center line and the speed limit of the road. $Q:=diag(0,c_y$, $c_\psi$, $c_v$, $c_d)$ where each scalar elements represent the tracking costs for $e_y^i$, $e_\psi^i$, $v^i$, and $d^i$, respectively. $\mathcal{R}_u$ $\mathcal{R}_{\Delta u}$ are the input cost matrix and input rate cost matrix, respectively. Note that these cost matrices are positive definite matrices and tuning parameters. (\ref{initial_cond}) sets the initial condition at time t/. $\mathcal{Z}$ is a feasible state set representing the road boundaries and vehicle dynamics limitations, $\mathcal{U}$ is a feasible input set, and $\Delta \mathbf{u}^{min}_{\cdot|\cdot}$ and $\Delta \mathbf{u}^{max}_{\cdot|\cdot}$ are input rate limits that arise from vehicle dynamic limitations. $z^i_{FSM}(t)$ is the FSM mode of the $i$-th CAV at time t at which the CFTOP (\ref{planner_cftoc}) is solved. The FSM mode is fixed during the horizon and its transition is described by (\ref{eq:FSM_dynamics}). 

As previously stated, the planner has three \textit{modes} with different constraints and tracking costs. Firstly, in \textit{LC} mode, a state terminal constraint (\ref{p_lc_terminal}) is applied. The terminal state set, $\mathcal{Z}_f$, is determined by the TL's lateral displacement from the OL and curvature at $s^i_{N|t}$ and requires the controlled vehicle to reach the target lane at the end of the horizon. (\ref{p_lc_const}) is applied to constrain the vehicle to the free space in the target lane and avoid collisions. The $s^{i,min}_{k|t}$ and $s^{i,max}_{k|t}$ are the longitudinal displacement of the rear and front (with respect to that of the controlled vehicle) vehicles in the TL at time $k$ predicted at time $t$, respectively, if any. The expectation-type constraint is applied because $s^{i,min}_{k|t}$ and $s^{i,max}_{k|t}$ may be stochastic NPC predictions. This constraint ensures that the controlled vehicle remains in free space in the TL while changing lanes within the horizon. Secondly, in \textit{LK} and \textit{GR} mode, (\ref{p_acc_const}) and (\ref{p_slack_const}) are applied for safe adaptive cruise control if there exists a preceding vehicle (its longitudinal position denoted as $s_{front,k|t}^i$) in the same lane as the controlled vehicle. Otherwise, (\ref{p_acc_const}) and (\ref{p_slack_const}) are not applied. To prevent infeasibility occurring from surrounding vehicles' sudden deceleration, the safety constraint (\ref{p_acc_const}) is relaxed by the time-varying slack variable $\epsilon^i_{k|t}$ with the cost $c_{s}$. In \textit{LC} mode, the desired $e_y$ is the lateral-displacement of the TL whereas it is zero (OL) in other modes. Lastly, the tracking cost for $d^i_{k|t}$ is only included in \textit{GR} mode. Accordingly, $c_d=q_d$ if in \textit{GR} mode and $c_d=0$ otherwise, where $q_d\in\mathbb{R}^+$ is a tracking cost for gap regulation and a design parameter.   

\subsection{Multi-CAV Conflict Prevention}

In distributed path planning of a multi-vehicle system, resolving trajectory conflicts between CAVs in the multi-CAV platoon is central to designing a safe and robust system. Trajectory conflicts in our lane change scenario may arise when lane change intention and motion predictions of other CAVs are not communicated and are wrong, respectively. For instance, during a simultaneous lane change of multiple CAVs, the controlled CAV's independently-computed optimal lane change trajectory may conflict with other CAVs' trajectories if their lane change intentions are not considered. Therefore, during a simultaneous lane change, a CAV must plan its trajectory while considering other lane changes of other CAVs. In our distributed coordination method, conflicts are prevented by using the following implementation steps.

Lane change intentions of other CAVs are considered when planning a simultaneous lane change of multiple CAVs by using the virtual vehicle-based method \citep{virtual_vehicle}. In this method, virtual vehicles are forecasts of CAVs onto their target lane and are considered when planning a lane change trajectory. The controlled CAV plans a lane change trajectory while considering the virtual CAVs in the target lane. This imposes stricter constraint (\ref{p_lc_const}) and reserve space for the virtual CAV's corresponding real CAV to change lanes into the free space, if feasible. Since all CAVs mutually consider other CAVs' virtual vehicles during simultaneous lane change, conflicts are prevented in their optimal lane change trajectories. 

Furthermore, in our planner formulation, the exact motion predictions of other CAVs at time \textit{t}, which are the state sequence $\mathbf{z}^i(\cdot|t)$ for $i\in\{1,2,...,n\}$ of CFTOP (\ref{planner_cftoc}), are obtained via V2V technology. Thus, the exact planned motion of other CAVs at time \textit{t} are known which allows the controlled CAV to plan a \textit{LC} trajectory that avoids collision with other CAVs as well as their virtual vehicles. During a simultaneous lane change of multi-CAV platoon, either the rear, front, or both vehicles are other CAVs' virtual vehicle. Therefore, at least one of the $s^{i,min}_{k|t}$ and $s^{i,max}_{k|t}$ terms in (\ref{p_lc_const}), which is the longitudinal motion predictions of rear and front vehicles in the target lane, is no longer an uncertain prediction but instead known exactly at time $t$ during simultaneous lane changes. 

\section{Numerical simulation} \label{numerical_sim}
We validate the proposed distributed coordination method and demonstrate the enhanced feasibility of platoon lane change using the \textit{proactive and cooperative} multi-CAV platoon lane change strategy in comparison to that of using \textit{passive and opportunistic} strategy (the baseline) through numerical simulations. In the baseline strategy, the multi-CAV will change lanes only when it is feasible for all three CAVs simultaneously. The nonlinear optimization problem (\ref{planner_cftoc}) is modeled using CasADi and solved using IPOPT \citep{casadi,ipopt}. In this section, the NPC controller and numerical simulation scenario setup are discussed. 

The following assumptions have been made in our numerical simulations:
1) the multi-CAV platoon has 3 CAVs with identical lengths, 2) the \textit{Facilitator} role is preassigned to one CAV. 

\subsection{NPC Controller}
To close the loop in our simulation, the NPC vehicles are controlled using the Enhanced Intelligent Driver Model (EIDM), which is based on the intelligent driver model (IDM). The EIDM relaxes the sudden deceleration in not safety-critical situations while inheriting IDM's crash-free characteristics \citep{enhancedIDM}. This model combines the IDM and constant-acceleration heuristics, which assumes that the leading vehicle will not change its acceleration for the next few seconds, to relax the sudden deceleration in not safety-critical situations \citep{enhancedIDM}. Furthermore, a scaling constant called \textit{coolness factor}, $c$, is a model parameter that determines the sensitivity with respect to changes of the gap with the vehicle in front. \cite{enhancedIDM} recommend $c\in[0.95,1.00]$ to simulate realistic driving behavior. This car-following model outputs vehicle acceleration to emulate an adaptive cruise control behavior. The EIDM parameters used in the simulation are reported in Table \ref{tab:parameters}. $s_0$ represents the minimum desired spacing between the controlled vehicle and the preceding vehicle. $v_0$ is the desired velocity, $T$ is the desired time headway, $a$ is the maximum vehicle acceleration, and $b$ is the comfortable braking deceleration where $b>0$.


\subsection{NGSIM Dataset Simulation} \label{ngsim_setup}
Our proposed planner was tested in a dense traffic scenario based on reconstructed NGSIM dataset \citep{ngsim_data}. The NGSIM dataset contains the recorded trajectories (sampling frequency of 10 Hz) of all vehicles passing a segment of freeway I-80 for 15 minutes \citep{ngsim_data}. The initial scene is generated by randomly selecting the initial frame and two adjacent lanes (i.e. lane 1 and 2) out of 5 total lanes. Then, three vehicles with the shortest inter-vehicle distances in the initial scene are selected as CAVs. The inter-vehicle distances must be smaller than the user-defined threshold which is set to 9 meters in our simulation. If the threshold condition is not satisfied, the scene is discarded and generated again randomly. After selecting the CAVs, the desired velocity of all CAVs and NPCs are set to the maximum velocity of all vehicles in the initial scene: $v^{max}_{scene}$. The NPC vehicles are simulated using the NPC controller as described previously. CAVs and NPCs are simulated in closed loop with kinematic bicycle model by applying the planner and NPC controller at 20 Hz update rate using the parameters in Table \ref{tab:parameters} for 25 seconds. Note that the NPC vehicles in the scene have different initial speed and time-varying acceleration, which renders it a more challenging scenario for multi-CAV platoon to change lanes. 
\begin{figure*}[t!]
    \centering \includegraphics[width=0.7\linewidth]{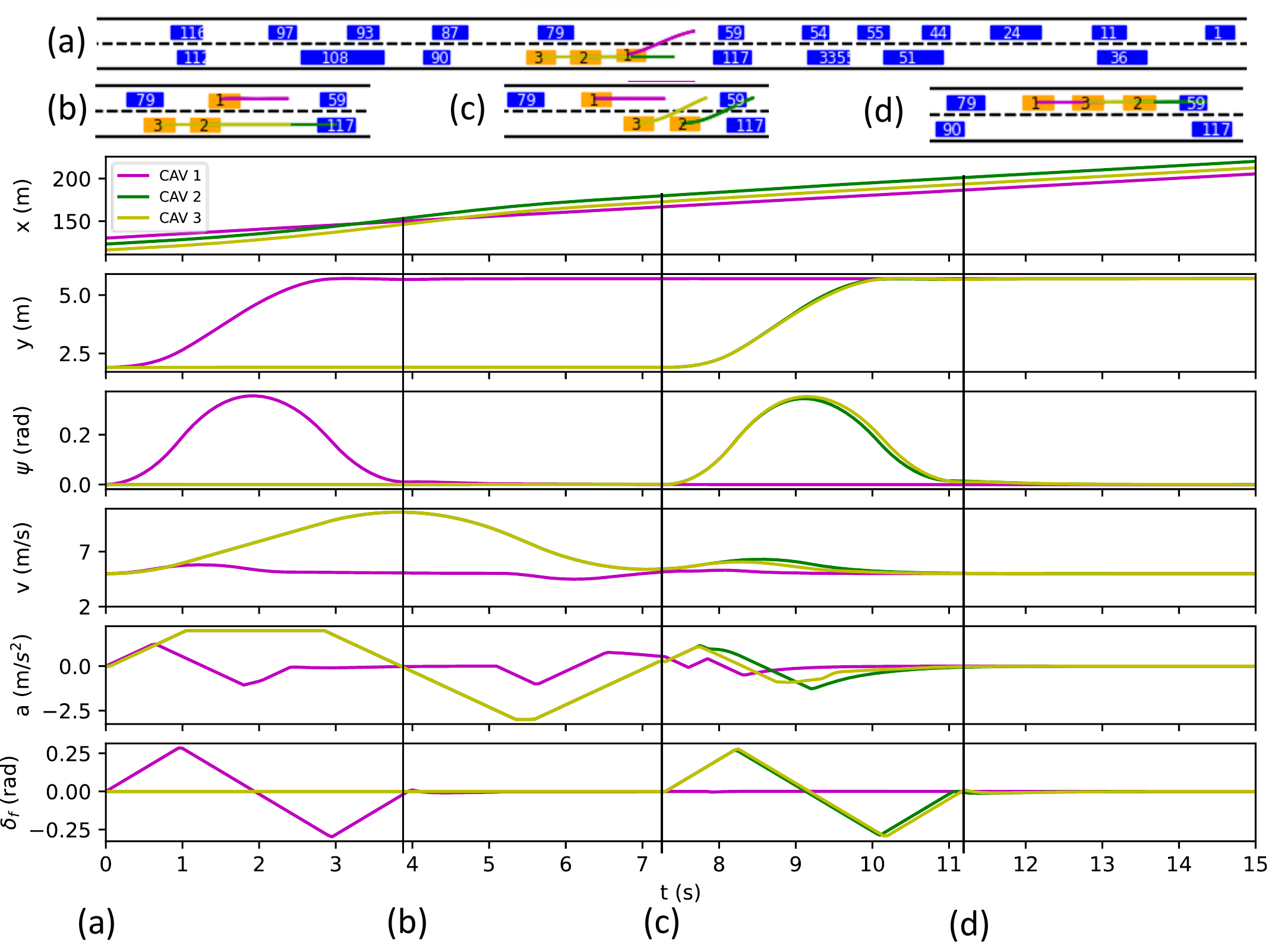}
    \caption{\textbf{Top:} A multi-CAV platoon changes lanes using the proactive and cooperative lane change strategy in a dense traffic scenario. Note that orange vehicle 1, vehicle 2, and vehicle 3 correspond to CAV 1, CAV 2, and CAV 3, respectively. Scene (a) shows the initial scene of lane 2 and 3 at frame 561 from the NGSIM dataset. Scene (b)-(c) show the facilitator's (CAV 1) changing lanes and regulating the distance while CAV 2 and 3 get in the desired formation. Scene (d) shows the completion of platoon-wise lane change. \textbf{Bottom:} The multi-CAV platoon vehicle's states and inputs are shown. The labeled sections on the time axis correspond to the scenes described above.   
    }
    \label{fig:statesandinputs}
\end{figure*}
\begin{table}
\caption{Numerical simulation parameters} \label{tab:parameters}
\centering
\begin{tabular}{cc|cc} 
\multicolumn{2}{c|}{\textbf{Vehicle Parameters }} & \multicolumn{2}{c}{\textbf{NPC Model Parameters}}  \\ 

\textbf{Parameter} & \textbf{Value}   & \textbf{Parameter}  & \textbf{Value} \\ 
\hline
$l_f$   & 2.235 m & $c$ & 1.0   \\ 

$l_r$      & 2.235 m  & $s_0$ & 2 m   \\ 

$w$  & 2 m    & $v_0$ & $v^{max}_{scene}$      \\ 

$v_{\text{min}}$  & 1.0 m/s &  $T$  & 1.0 s    \\ 

$v_{\text{max}}$   & 32 m/s & $a$ & 0.73 $\text{m/s}^2$   \\ 

$a_{\text{min}}$ & -3 $\text{m/s}^2$ & $b$ & 1.67 $\text{m/s}^2$  \\ 

$a_{\text{max}}$ & 2 $\text{m/s}^2$& $\sigma^2$ & 0.2 $\text{m/s}^2$\\ \cline{3-4}
$\delta_{\text{min}}$ & -0.4 & \multicolumn{2}{c}{\textbf{Planner Parameters}} \\
 $\delta_{\text{max}}$& 0.4  & \textbf{Parameter} & \textbf{Value}\\ \cline{3-4}
 $\Delta a_{\text{min}}$& -2 $\text{m/s}^3$ & N & 40 \\
 $\Delta a_{\text{max}}$& 2 $\text{m/s}^3$ & $\Delta t$ & 0.05 s  \\
 $\Delta \delta_{\text{min}}$& -0.3 rad/s & $d_{safe}$ &  $2 + (l_f+l_r)$ m \\
 $\Delta \delta_{\text{max}}$ & 0.3 rad/s& $n$ & 3 \\ \cline{1-2}
\multicolumn{2}{c|}{\textbf{Scene Parameter}} & $c_{s}$ & 20 \\
\textbf{Parameter} & \textbf{Value} & $c_{y}, c_{\psi}, q_{d}$  & 3  \\ \cline{1-2}
Lane Width& 3.8 m & $c_{v}$   & 2  \\
\hline
\end{tabular}
\end{table}

In 200 randomly generated dense traffic scenarios, the baseline and our proposed lane change strategies were applied. The performance metrics are the number of successful lane change completion of all three CAVs and the average lane change competition time in successful lane change simulation runs. Table \ref{tab:ngsim_results} reports the obtained results. Over two-fold increase in number of lane change completions using our proposed strategy is mainly attributed to the single vehicle lane change characteristics. It only requires a single vehicle (facilitator) to change lanes to initiate the coordinated maneuvers to enable multi-CAV lane change whereas the baseline controller requires three-vehicles to change lanes simultaneously. After changing lanes, the facilitator regulates the free space by proactively modifying the environment to render the lane change feasible for the other CAVs. Moreover, multi-CAV platoon completed lane change using the proposed strategy in every traffic scenario that the baseline strategy successfully completed lane change. On average, the baseline strategy takes 12.20 seconds to complete a multi-CAV platoon lane change while the proposed strategy takes 12.64 seconds. Fig. \ref{fig:statesandinputs} shows snapshots of platoon-wise lane change using the proposed strategy in a dense traffic scenario generated from NGSIM dataset as well as time series of states and inputs of the CAVs during the simulation. In this scenario, platoon-wise lane change using the baseline strategy is not feasible during the duration of the simulation.

\begin{table}
\caption{Performance comparison of baseline and proposed lane change strategies} \label{tab:ngsim_results}
\centering
\begin{tabular}{ccc} 
\textbf{} & Num. of $LC^P_{complete}$  & Avg. $LC^P_{complete}$ time \\ 
\hline
\textbf{Baseline}&27 & 12.20 s \\ 
\textbf{Proposed} & 57 & 12.64 s \\ 
\hline
\end{tabular}
\end{table}

\section{Conclusion} \label{Conclusion}

In this paper, we presented the concept of a facilitator and a proactive and cooperative multi-CAV platoon lane change strategy. A distributed coordination method with MPC path-planners with three modes and a higher-level FSM to manage mode transitions were used to implement the proposed lane change strategy. Leveraging vehicle connectivity, the facilitator assists the multi-CAV platoon's objective by proactively modifying the environment and enhances, as a result, the feasibility of the lane change objective as shown in numerical simulations. Tested in 200 dense traffic scenarios randomly generated based on the NGSIM dataset, the proposed lane change strategy demonstrated over a two-fold increase in numbers of lane change completion of a three-CAV platoon in comparison to the \textit{passive and opportunistic} lane change strategy. 
The simulation results substantiate the utility and potential of proactive and cooperative approach to multi-CAV lane change in certain traffic conditions (i.e. dense traffic and small speed difference between OL and TL). The future work will consider vehicle interaction model and a facilitator selection algorithm to improve the performance. Experimental results are also envisaged for the future.

\bibliography{ifacconf}  

\end{document}